\documentclass[runningheads]{llncs}
\usepackage[utf8]{inputenc} 
\usepackage[T1]{fontenc}    
\usepackage{url}            
\usepackage{booktabs}       
\usepackage{nicefrac}       
\usepackage{microtype}      
\usepackage{xcolor}         

\usepackage{amsmath}
\usepackage{amsfonts}
\usepackage{amssymb}

\usepackage{float}
\usepackage{graphics}
\usepackage{graphicx}
\usepackage{caption}
\usepackage{algorithm}
\usepackage{algpseudocode}
\usepackage{bm}
\usepackage{color}
\usepackage{multirow}
\usepackage{makecell}
\usepackage{balance}
\usepackage{diagbox}
\usepackage{subfigure}
\usepackage{enumitem}
\usepackage{apptools}

\usepackage[title]{appendix}

\allowdisplaybreaks

\AtAppendix{\counterwithin{thm}{section}}
\AtAppendix{\counterwithin{lem}{section}}
\AtAppendix{\counterwithin{defi}{section}}
\AtAppendix{\counterwithin{prop}{section}}
\AtAppendix{\counterwithin{rmk}{section}}
\AtAppendix{\counterwithin{assumption}{section}}

\algnewcommand\algorithmicforpara{\textbf{for}}
\algnewcommand\algorithmicdoinparallel{\textbf{do in parallel}}
\algdef{S}[FOR]{ForParallel}[1]{\algorithmicforpara\ #1\ \algorithmicdoinparallel}

\begin{document}
\title{Federated Co-tuning Framework for Large and Small Language Models}

\titlerunning{Federated
Co-tuning Framework for Large and Small
Language Models}

\author{Tao Fan\inst{1,2}\and
Yan Kang\inst{2}\and
Guoqiang Ma\inst{2}\and
Lixin Fan\inst{2} \and \\
Shuoling Liu \inst{1} \and
Kai Chen\inst{1} \and
Qiang Yang\inst{3}}

\institute{
Hong Kong University of Science and Technology, Hong Kong, China\\ 
 \and
WeBank Co., Ltd, Shenzhen, China \\
\and
Hong Kong Polytechnic University, Hong Kong, China
\\ 
{\{tfanac, sliudi, kaichen, qyang\}@cse.ust.hk} 
\\{\{yangkang, zotrseeewma, lixinfan\}@webank.com}
}

\maketitle              

\begin{abstract}

By adapting Large Language Models (LLMs) to domain-specific tasks or enriching them with domain-specific knowledge, we can fully harness the capabilities of LLMs. Nonetheless, a gap persists in achieving simultaneous mutual enhancement between the server's LLM and the downstream clients' Small Language Models (SLMs). To address this, we propose FedCoLLM, a novel and parameter-efficient federated framework designed for co-tuning LLMs and SLMs. This approach is aimed at adaptively transferring server-side LLMs knowledge to clients' SLMs while simultaneously enriching the LLMs with domain insights from the clients. To accomplish this, FedCoLLM utilizes lightweight adapters in conjunction with SLMs, facilitating knowledge exchange between server and clients in a manner that respects data privacy while also minimizing computational and communication overhead. Our evaluation of FedCoLLM, utilizing various public LLMs and SLMs across a range of NLP text generation tasks, reveals that the performance of clients' SLMs experiences notable improvements with the assistance of the LLMs. Simultaneously, the LLMs enhanced via FedCoLLM achieves comparable performance to that obtained through direct fine-tuning on clients' data. Our code has been contributed to the FATE open-source project and is now publicly accessible at \url{https://github.com/FederatedAI/FATE-LLM/tree/main/python/fate_llm/algo/fedcollm}.


\end{abstract}

\section{Introduction} 
The emergence of Large Language Models (LLMs) has profoundly transformed the landscape of artificial intelligence. In particular, cutting-edge LLMs like GPT-4~\cite{Gpt-4} have garnered significant attention due to their exceptional performance across a range of natural language generation tasks. This development has spurred the release of numerous high-performance open-source LLMs, such as LLaMa~\cite{touvron2023llama}, OPT~\cite{zhang2022opt}, greatly promoting the commercial application of LLMs technology.
Despite their widespread success in various general NLP tasks, LLMs face limitations that hinder their adoption in domain-specific applications. The primary challenges include:

\begin{itemize}

\item \textbf{Domain-Specific Knowledge Privacy}. 
When downstream clients are unable to access the LLMs parameter, they have to send their labeled data to the LLMs owners for fine-tuning. This process inevitably discloses the privacy of clients' sensitive domain-specific data.

\item \textbf{Constrained Resources}.
Even when downstream enterprises can obtain the model parameters of LLMs, they often encounter significant resource constraints. Fine-tuning these LLMs requires substantial computing and storage resources, posing a barrier to adoption by small and medium-sized companies with limited resources. As a result, these companies are restricted to fine-tuning Small Language Models(SLMs) using their domain-specific data.

\item \textbf{Mutual Knowledge Transfer Between LLMs and SLMs}. Optimizing LLMs and SLMs separately forms a positive feedback loop for continuous evolution. Server-side LLMs first transfer general knowledge to client-side SLMs, which are then fine-tuned on domain-specific data for downstream tasks. These domain-specialized SLMs can in turn feed industry-specific knowledge back to server LLMs, improving their understanding and expanding their scope and depth. However, such mutual enhancement between server LLMs and client SLMs remains largely unexplored in existing literature.

\end{itemize}

To address the aforementioned challenges, we propose FedCoLLM, an innovative and parameter-efficient federated co-tuning framework for LLMs and SLMs. This framework is designed to enhance the performance of both server-side LLMs and client-side SLMs.
As illustrated in Figure \ref{fig:fedcollm}, FedCoLLM deploys a LLM on the server and introduces a SLM to serve as a bridge between the privacy data in the clients and LLM in the server. The SLM operates simultaneously across multiple clients and the server, facilitating efficient communication and collaboration.
FedCoLLM offers three distinct advantages:
\begin{itemize}

\item \textbf{Efficient Computation and Communication}. 
FedCoLLM initially runs the SLM under standard federated learning(FL) frameworks ~\cite{mcmahan2017communication,yang2019federated}. This approach integrates a parameter-efficient adapter module, such as LoRA~\cite{hu2021lora}, significantly reducing the computation and communication costs associated with FedCoLLM.

\item  \textbf{Enhanced Data Privacy}.
By leveraging the FL framework to fine-tune the SLM, FedCoLLM fully utilizes the FL security protection mechanisms(such as SecureAggregation~\cite{bonawitz2016practical}) to preserve the privacy of clients' data. This ensures that sensitive information remains protected during the fine-tuning process. 

\item \textbf{Knowledge Transfer and Mutual Enhancement}. 
FedCoLLM employs knowledge distillation(KD) techniques~\cite{hinton2015distilling}, to transfer knowledge between the LLM and SLM on the server. This process is facilitated by an auxiliary distillation dataset, making it particularly beneficial for clients with limited resources. Through this knowledge transfer, both the server LLM and client SLMs are mutually enhanced, leading to improved overall performance.

\end{itemize}
Extensive experiments conducted on various LLMs and SLMs, including GPT-2~\cite{radford2019language}, OPT ~\cite{zhang2022opt}, and LLaMa2~\cite{touvron2023llama}, demonstrate the competitive performance of our FedCoLLM framework across a range of NLP text generation tasks. The results show that the SLMs can achieve significant enhancements with the support of the LLM, while the LLM can deliver comparable results to fine-tuning with all clients' domain data directly. Importantly, our framework is more resource-efficient, requiring lower computation and communication costs.

\section{Related Work}

\subsection{Knowledge Distillation}
Knowledge distillation is a technique that has gained significant attention in recent years, as it enables the transfer of knowledge from a larger teacher model to a smaller student model. One of the early works in this area was proposed by~\cite{hinton2015distilling}, which introduced the concept of knowledge distillation and demonstrated its effectiveness in improving the performance of compressed models. Since then, numerous studies have built upon this foundation and explored various distillation strategies~\cite{gou2021knowledge}. For instance, two notable work ~\cite{chen2017learning,meng2019conditional} improved response-based knowledge distillation, which let the student model directly mimic the final prediction of the teacher model. FitNets~\cite{adriana2015fitnets} focus on matching intermediate representations between the teacher and student models. 
Another notable work is the relational knowledge distillation approach introduced by~\cite{park2019relational}, which captures pairwise relationships between outputs to enhance distillation efficiency. 
Different from one-way knowledge distillation between the teacher and student networks, Deep Mutual Learning (DML)~\cite{zhang2018deep} allows two networks to learn from each other through their predicted probability distributions during the training process.
These studies have demonstrated the potential of knowledge distillation in various tasks, such as image classification, object detection, and natural language processing.

\subsection{Federated Learning for Large Language Models}

Parameter-Efficient Fine-Tuning (PEFT) techniques~\cite{hu2021lora} provide a straightforward remedy to address the challenges of communication overhead and fine-tuning expenses in Federated Learning (FL) for Large Language Models (LLMs)~\cite{fan2023fate,kang2023grounding,fan2025ten}. Numerous investigations have extended the application of PEFT methods within the FL framework tailored for LLMs. Notable contributions include FedPETuning~\cite{zhang2023fedpetuning}, Federated Adapter Tuning~\cite{cai2023efficient}, and Federated Prompt Tuning~\cite{zhao2023fedprompt}.
These research outcomes suggest that FL clients, particularly those with constrained storage capacities like mobile devices, can significantly profit from the adoption of PEFT methods. These approaches facilitate the sharing of LLMs across various tasks while necessitating the retention of only a minimal set of parameters per task, effectively decreasing storage demands. Through the utilization of PEFT methods, FL clients can adeptly tailor LLMs to meet their unique requirements, all the while minimizing communication overhead and reducing fine-tuning costs.

\section{The Proposed Method}
In this section, we present a comprehensive overview of our federated co-tuning LLMs and SLMs framework, termed FedCoLLM. This framework is based on parameter-efficient fine-tuning (PEFT) and knowledge distillation techniques. We begin by defining the specific problem addressed in this study, followed by a detailed introduction to our approach. Finally, we delve into the computational and communication complexities, as well as the privacy-preserving analysis, of our FedCoLLM framework.

\subsection{Problem Definition}
In this work, we consider the federated learning setting in which the server owns an LLM $f_\psi$ parameterized by $\psi$ and $K$ clients that each client $k$ has a SLM $g_\phi$ parameterized by $\phi$. The server and clients aim to collaboratively enhance the performance of the LLM and SLMs without sharing private data through federated learning. Specifically, 
\begin{itemize}
    \item Each client possesses its own local private dataset $\mathcal{D}^k$. Clients aim to collectively train a global SLM $g_{\phi}$ based on their local models initialized with an SLM (e.g., LLaMa2-1.3B~\cite{xia2023sheared}) without divulging their private data. The objective can be formulated as follows:
     \begin{equation}
           \min_\phi \mathcal{L}_1(\phi;\{\mathcal{D}^k\}_{k=1}^K) \label{eq:hfl}
    \end{equation}

    \item The server owns an auxiliary dataset $D^a$. The server aims to transfer knowledge between its owned LLM $f_{\psi}$ and the global SLM $g_{\phi}$ aggregated from clients' local SLMs to enhance both the LLM and SLMs. The objective can be formulated as follows:
    \begin{equation}
        \min_{\phi,\psi} \mathcal{L}_2(\phi, \psi; \mathcal D^a) \label{eq:mkt}
    \end{equation}
\end{itemize}
We regard the server as semi-honest. FedCoLLM solves the optimization problems formulated in Eq.(\ref{eq:hfl}) and Eq.(\ref{eq:mkt}) in an efficient and privacy-preserving manner. We will elaborate on FedCoLLM in Section \ref{sec:fedcollm-spec}.

\subsection{FedCoLLM}
\label{sec:fedcollm-spec}
FedCoLLM is a novel framework designed to facilitate the collaborative evolution of both server-side LLM and client-side SLMs. The goal of the FedCoLLM is threefold: 
\begin{itemize}
\item \textbf{Collaborative Knowledge Transfer and Adaptation}. 
The server and clients work together to transfer and adapt the knowledge of the LLM owned by the server. This helps clients build local SLMs that benefit from the server's LLM knowledge. By leveraging the server's LLM, clients can improve their local SLMs' performance without requiring extensive local training data or computational resources.

\item \textbf{Data Augmentation for the Server's LLM}. 
Federated learning also aims to leverage clients' data to augment and enhance the server's LLM. Clients' data often contains valuable local information and patterns that can be used to improve the server model's generalization and performance. By incorporating this data, the server's LLM can become more robust and adaptive to different scenarios and domains.

\item \textbf{Privacy-Preserving and Efficient Knowledge Transfer}. 
A crucial aspect of federated learning is ensuring that knowledge transfer occurs in a privacy-preserving and efficient manner. Clients' raw private data should not be directly uploaded to the LLM server, preserving their privacy. Instead, only model updates or aggregated information are shared with the LLM server. Additionally, the knowledge transfer process should be efficient, minimizing communication costs and computational overhead.

\end{itemize}

Toward this goal, we (1) adopt lightweight LoRA modules as the bridge to transfer the knowledge between clients and the server, (2) leverage mutual knowledge distillation to transfer knowledge between the LLM and the aggregated SLM, and (3) employ secure aggregation to protect the privacy of the knowledge transfer process. 
Specifically, we assume that clients and the server share a SLM $g_{\phi}$ parameterized by $\phi$. Each client $k$ inserts a small low-rank adapter parameterized by $\theta_k$ into its local SLM. We denote a client's local SLM with the added $\theta$ as $g_{\phi+\theta}$. Instead of training a global SLM $\phi$, clients collaboratively train a global LoRA module $\theta$. Then, Eq.(\ref{eq:hfl}) can be reformulated as follows:
\begin{equation}\label{eq:fedcollm}
\begin{aligned}
     \mathcal{L}_1(\theta;\{\mathcal{D}^k\}_{k=1}^K) = &\frac{1}{K}\sum_{k=1}^K \mathbb{E}_{(x,y) \sim \mathcal{D}^k} \ell_{\text{TA}}^k(g_{\phi+\theta}(x), y).
\end{aligned}
\end{equation}
where $\ell_{\text{TA}}$ is the task loss for training the global LoRA module $\theta$. The original model parameter $\phi$ of each client's local SLM is frozen during training.

The server inserts a small low-rank adapter parameterized by $\omega$ into its LLM $f_{\psi}$. We denote the server's LLM $f_{\psi}$ with the added $\omega$ as $f_{\psi+\omega}$. The server conducts the mutual knowledge transfer between the LLM $f_{\psi+\omega}$ and the global SLM $g_{\phi+\theta}$ through supervised fine-tuning and mutual knowledge distillation based on the auxiliary dataset $\mathcal{D}^a$.
We formulate the losses of supervised fine-tuning $f_{\psi+\omega}$ and $g_{\phi+\theta}$ (denoted as $\mathcal{L}_{\text{FT}}^f$
and $\mathcal{L}_{\text{FT}}^g$) as follows:

\begin{equation}\label{eq:mft}
\begin{aligned}
     \mathcal{L}^f_{\text{FT}}(\omega;\mathcal{D}^a)= \mathbb{E}_{(x,y) \sim \mathcal{D}^a}\ell_{\text{CE}}(f_{\psi+\omega}(x), y), \\
     \mathcal{L}^g_{\text{FT}}(\theta;\mathcal{D}^a)= \mathbb{E}_{(x,y) \sim \mathcal{D}^a} \ell_{\text{CE}}(g_{\phi+\theta}(x),y).
\end{aligned}
\end{equation}
where $\ell_{\text{CE}}$ is the cross-entropy loss; the model parameters $\psi$ and $\phi$ are frozen during fine-tuning. 

The mutual knowledge distillation losses for fine-tuning $f_{\psi+\omega}$ and $g_{\phi+\theta}$ models (denoted as $\mathcal{L}_{\text{KD}}^f$
and $\mathcal{L}_{\text{KD}}^g$) are formulated as follows:
\begin{equation}\label{eq:mkd}
\begin{aligned}
     \mathcal{L}^f_{\text{KD}}(\omega;\mathcal{D}^a)= \mathbb{E}_{x \sim \mathcal{D}^a}\ell_{\text{KL}}(f_{\psi+\omega}(x), g_{\phi+\theta}(x)),  \\
     \mathcal{L}^g_{\text{KD}}(\theta;\mathcal{D}^a)= \mathbb{E}_{x \sim \mathcal{D}^a} \ell_{\text{KL}}(g_{\phi+\theta}(x),f_{\psi+\omega}(x)).
\end{aligned}
\end{equation}
where $\ell_{\text{KL}}$ is the Kullback Leibler (KL) divergence function; the model parameters $\psi$ and $\phi$ are frozen during knowledge distillation. 

Combining Eq.(\ref{eq:mft}) and Eq.(\ref{eq:mkd}), we formulate the mutual knowledge transfer conducted on the server as follows:
\begin{equation}\label{eq:fedcollm_2}
\begin{aligned}
   & \mathcal{L}_2(\theta, \omega; \mathcal D^a) = \mathcal{L}^f(\omega;\mathcal{D}^a) + \mathcal{L}^g(\theta;\mathcal{D}^a),\\
   & \text{in which}\\
   & \mathcal{L}^f(\omega;\mathcal{D}^a) =  \mathcal{L}_{\text{FT}}^f(\omega;\mathcal{D}^a) + \lambda\mathcal{L}_{\text{KD}}^f(\omega;\mathcal{D}^a), \\
   &   \mathcal{L}^g(\theta;\mathcal{D}^a) =  \mathcal{L}_{\text{FT}}^g(\theta;\mathcal{D}^a) + \lambda \mathcal{L}_{\text{KD}}^g(\theta;\mathcal{D}^a).\\
\end{aligned}
\end{equation}
where $\lambda$ is the hyperparameter that controls the weight of mutual knowledge transfer.

After mutual knowledge transfer, the global LoRA module $\theta$ is distributed to all clients, which in turn adopts Eq.(\ref{eq:hfl}) to further train $\theta$ based on their local datasets.

\begin{figure}[!h]
  \centering
  \includegraphics[width=.60\textwidth]{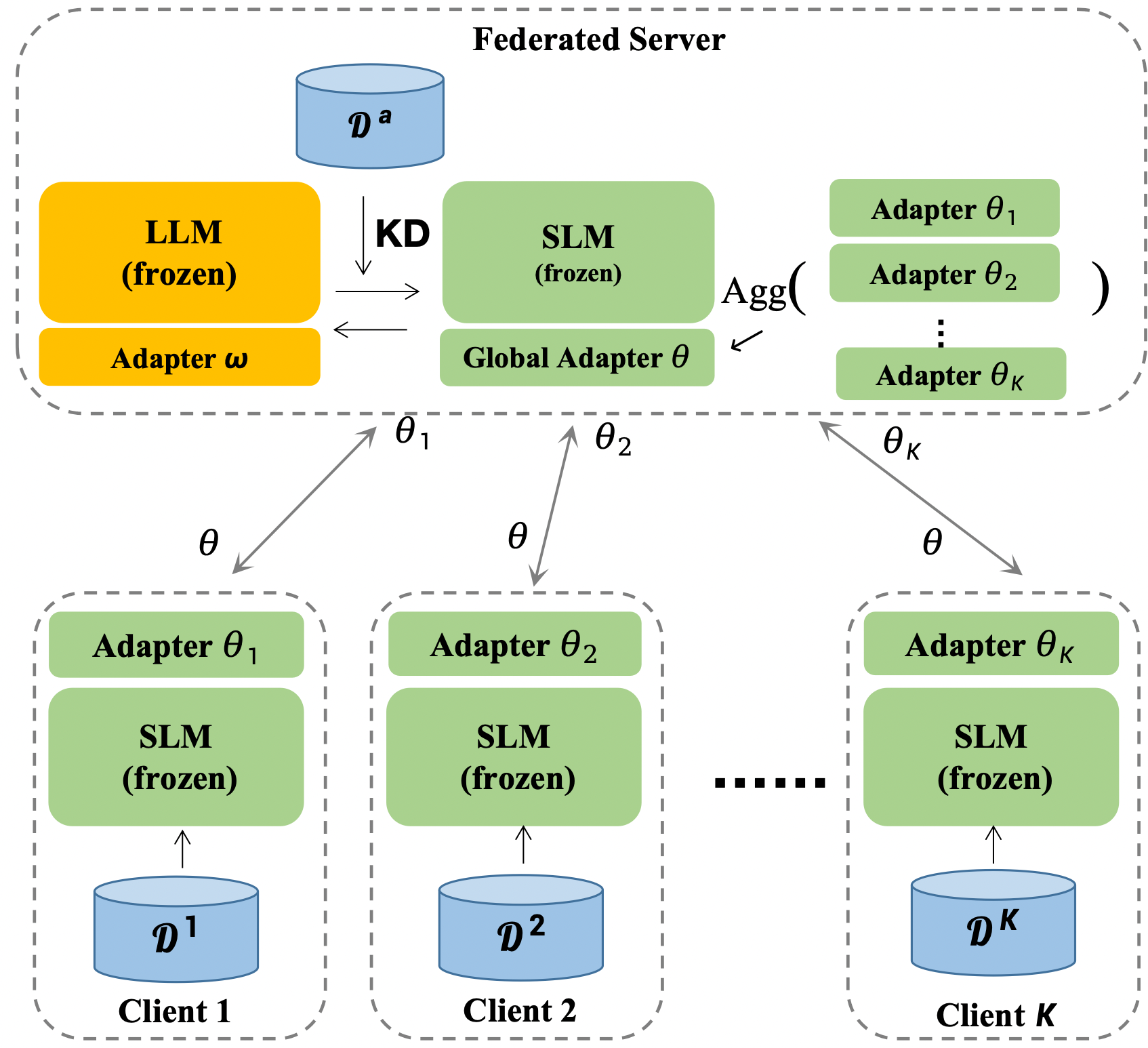}
  \caption{\textbf{FedCoLLM} (Federated parameter-efficient co-tuning of clients’ domain SLMs and the server’s LLMs. Clients’ SLMs learn from each other via federated fine-tuning of their adapter modules and transfer knowledge from and to the server’s LLM)}
  \label{fig:fedcollm}
\end{figure}

FedCoLLM thus fosters a symbiotic relationship between the server and clients, where both parties benefit from the collective knowledge and expertise encoded in their respective language models. By leveraging the complementary strengths of server-side LLMs and client-side SLMs, FedCoLLM paves the way for more efficient and effective federated learning in the realm of natural language processing, enabling the collaborative evolution of LLM and SLMs.  We illustrate the FedCoLLM in Figure~\ref{fig:fedcollm} and describe the associated training algorithm in Algorithm \ref{alg:FedCoLLM}. The workflow of FedCoLLM proceeds as follows:
\begin{enumerate}

\item In the $t$-th communication round, the server broadcasts the SLM $g_{\phi+\theta}$ global adapter $\theta$ to $K$ clients. Each client $k$ then replaces its local adapter $\theta_k$ with the received global adapter $\theta$.

\item During local training, the $K$ clients fine-tune their respective local adapters using their private data. This step allows the clients to adapt their models to their specific data distributions while preserving the knowledge encoded in the global adapter.

\item After local training, the $K$ clients send their respective local adapters to the server. The server SLM $g_{\phi+\theta}$ then aggregates these local adapters using a secure averaging technique, such as SecureAvg, and updates the global adapter $\theta$ in the SLM accordingly.

\item  On the server side, LLM $f_{\psi+\omega}$ and SLM $g_{\phi+\theta}$ engage in knowledge distillation. This process involves transferring knowledge between the two models with the aid of an auxiliary distillation dataset. Through this distillation, both models can benefit from each other's learned representations, leading to improved performance and adaptability.

\end{enumerate}

\subsection{Computation and Communication Complexity}

One key advantage of FedCoLLM is its computational efficiency. By utilizing PEFT, it markedly decreases the parameters needing fine-tuning updates. Furthermore, the server-side distillation process compresses knowledge from local models into a smaller global model, optimizing computational resources. This enables effective learning from all clients' collective data while keeping the model size manageable.
In terms of communication complexity, FedCoLLM minimizes the amount of data exchanged between clients and the server. Instead of transmitting entire models or large datasets, clients only share their locally fine-tuned model updates with the server. This approach significantly reduces communication overhead.

\subsection{Privacy-Preserving Analysis}

FedCoLLM is meticulously designed with privacy preservation as its foundation. Recognizing the importance of data confidentiality, the framework ensures clients never directly disclose raw local data. Privacy protection is further enhanced through PEFT and knowledge distillation, minimizing sensitive information exposure during training.
By using knowledge distillation, FedCoLLM transfers key insights to a unified global model, sharing only aggregated, non-sensitive knowledge and preserving individual client privacy. Additionally, it seamlessly integrates with the standard FL framework for SLMs fine-tuning, leveraging security mechanisms like SecureAggregation~\cite{bonawitz2016practical} to maintain robust data privacy for all clients.

\begin{algorithm}[h] 
\caption{FedCoLLM}
\label{alg:FedCoLLM}
\scriptsize 
\begin{algorithmic}[1]
\renewcommand{\algorithmicrequire}{\textbf{Input:}}
\renewcommand{\algorithmicensure}{\textbf{Output:}}

\Require  \\
    $K$: number of clients;\\
    $T$: total number of rounds; \\
    $\mathcal{D}^{a}$: the auxiliary distillation dataset; \\
    $\mathcal{D}^{k}$: local datasets for each client $k$; \\
    $\eta_\omega$: the learning rate for optimizing LLM $f_{\psi+\omega}$; \\
    $\eta_\theta$: the learning rate for optimizing SLM $g_{\phi+\theta}$.
    
\Ensure $f_{\psi+\omega}$, $g_{\phi+\theta}$.
    \For{$t = 1, 2, \dots, T$}
        \State $// \textbf{ Server side:}$
        
        \State Broadcast the global adapter $\theta^{t}$ to all $K$ clients.
        
        \State $\theta_{k}^{t+1} \gets \textbf{ClientUpdate}(\theta^{t})$ for each client $k$
        
        \State Initialize the global adapter $\theta^{t,0} \gets \frac{1}{K} \sum_{k=1}^{K} \theta_{k}^{t+1}$ 

        \State $\triangleright$ mutual knowledge transfer based on $\mathcal{D}^{a}$

        \For{$r = 1, 2, \dots, R$} 
        
            \State $\theta^{t,r+1} \gets \theta^{t,r} - \eta_\theta \nabla \mathcal{L}^g $

            \State $\omega^{t,r+1} \gets \omega^{t,r} - \eta_\omega \nabla \mathcal{L}^f $
            
        \EndFor
        
        \State $\theta^{t+1} \gets \theta^{t,R} $
        \State $\omega^{t+1} \gets \omega^{t,R} $
        
        \State
        
        \Function{ClientUpdate}{$\theta^{t}$} 
            \State Receive the global adapter $\theta^{t}$ from the server;
            \For{each client $k$ (in parallel)}
                \State $\triangleright$ local fine-tuning based on $\mathcal{D}^{k}$
                \State $\theta_{k}^{0} \gets \theta^{t}$
                \For{$e = 1, 2, \dots, E$} 
                    \State $\theta_{k}^{e+1} \gets \theta_{k}^{e} - \eta_\theta \nabla \ell_{\text{TA}}^k$
                \EndFor
                \State $\theta_{k}^{t+1} \gets \theta_{k}^{E}$
                \State Upload updated local adapter $\theta_{k}^{t+1}$ to the server.
            \EndFor
            \State \Return $\theta_{k}^{t+1}$
        \EndFunction
        
    \EndFor 
    
\end{algorithmic}
\end{algorithm}

\section{Experiments}

\subsection{Setup}
We set up a scenario involving four clients and one server to evaluate the FedCoLLM using various LLMs and SLMs.

\textbf{Models}. We evaluate FedCoLLM on LLMs and SLMs, including GPT-2~\cite{radford2019language}, OPT~\cite{zhang2022opt} and LLaMa2~\cite{touvron2023llama}.
Our experiments involve utilizing the FedCoLLM framework with LLMs and SLMs of identical architecture but different model sizes. Specifically, for example, we employ LLaMa2-7B as the LLM and LLaMa2-1.3B~\cite{xia2023sheared} as the SLM in the FedCoLLM framework. 

\textbf{Datasets}. We evaluate FedCoLLM on 4 QA datasets, including CommonsenseQA(CQA)~\cite{talmor2018commonsenseqa}, OpenBookQA~\cite{mihaylov2018can}, ARC-C~\cite{clark2018think}, ARC-E~\cite{clark2018think}. 

\textbf{Baselines}. 
We conducted a comparative analysis of our FedCoLLM framework against several baselines to evaluate its performance. These baselines included:
\begin{itemize}
    \item Zero-Shot, which represents the zero-shot capabilities of the of LLM or SLMs.
    \item Standalone, where each client independently fine-tunes its local model using its own private dataset;
    \item FedAvg, in which clients train on their private datasets using the FedAvg algorithm\cite{mcmahan2017communication}; 
    \item Centralized, where the server's LLM is fine-tuned locally using the entirety of the private datasets combined with an auxiliary distillation dataset. 
\end{itemize}

\textbf{Evaluation Metrics}. 
We evaluate the model performance of fine-tuned LLMs and SLMs on the QA datasets using Accuracy as the primary metric. It's worth noting that in our experiments, all methods undergo zero-shot evaluation, and we use the \textit{lm-evaluation-harness} package. Additionally, to assess the communication efficiency of our framework, we measure the communication cost by tracking the number of transmitted parameters. 

\subsection{Performance Evaluations}

We conduct experiments with three settings. The first setting (denoted as S1) involves one server-side GPT-2-Large LLM and four client-side GPT-2-Small SLMs,  the second setting (denoted as S2) involves one server-side OPT-6.7B LLM and four client-side OPT-1.3B SLMs, and the third setting (denoted as S3) involves one server-side LLaMa2-7B LLM and four client-side LLaMa2-1.3B SLMs.
Tables \ref{tab:Homogeneous-LLM},\ref{tab:Homogeneous-SLM} demonstrate the performance comparisons of our approach against other baselines. The top sub-table and the bottom sub-table compare the performance of FedCoLLM against baselines on the server's LLM and clients' SLMs, respectively.

The Table \ref{tab:Homogeneous-LLM} shows that FedCoLLM significantly outperforms Zero-Shot on the server's LLM in the three settings. It also shows that FedCoLLM achieves comparable performance of the Centralized scenario. For example, in the CQA dataset, FedCoLLM achieves a relative improvement of 47\% over Zero-Shot on GPT-2-Large LLM, 40\% on OPT-6.7B LLM, and 70\% on LLaMa2-7B LLM. Furthermore, FedCoLLM nearly equals Centralized performance, reaching 98\% on GPT-2-Large LLM, 99\% on OPT-6.7B LLM, and 97\% on LLaMa2-7B LLM.

The Table \ref{tab:Homogeneous-SLM} shows that the SLM of FedCoLLM performs better than the Zero-Shot, Standalone, and FedAvg due to the assistance of the server's LLM. 
For example, in the CQA dataset, FedCoLLM achieves a relative improvement of 6\% over Standalone on GPT-2-Small SLM, 6\% on OPT-1.3B SLM, and 3\% on LLaMa2-1.3B SLM. Furthermore, FedCoLLM achieves a relative improvement of 3\% over FedAvg on GPT-2-Small SLM, 4\% on OPT-1.3B SLM, and 2\% on LLaMa2-1.3B SLM.

\begin{table}[!ht]
\centering
\footnotesize
\setlength{\tabcolsep}{2.8pt}

\begin{tabular}{c|c|c|c|c}
\hline
 \hline
\multirow{2}{*}{\shortstack{\textbf{Task}}} & \multirow{2}{*}{\shortstack{{\textbf{Method}}}} & 
\multirow{2}{*}{\shortstack{S1: Server \\ \textbf{GPT-2-Large}}} & 
\multirow{2}{*}{\shortstack{S2: Server \\\textbf{OPT-6.7B}}}&
\multirow{2}{*}{\shortstack{S3: Server \\ \textbf{LLaMa2-7B}}}\\
~ & ~ & ~ & ~ & ~\\  

\hline

\multirow{3}*{CQA}& Zero-Shot& 36.3 & 48.7 & 39.5\\
\cline{2-5}
~& Centralized& 54.7 & 68.6 & 69.0\\
 \cline{2-5}
~& \textbf{FedCoLLM}& 53.5 & 68.1 &67.1\\ 

 \hline

 \multirow{3}*{OBQA}& Zero-Shot& 19.4 & 27.6 & 31.8 \\
\cline{2-5}
~& Centralized&28.2 &34 & 39.8 \\
 \cline{2-5}
~& \textbf{FedCoLLM}& 25.4 & 34.4 & 37.8\\ 

 \hline

\multirow{3}*{ARC-C}& Zero-Shot& 21.7 & 30.7 &40.0\\

\cline{2-5}
~& Centralized& 28.8 & 37.1 &49.0\\
 \cline{2-5}
~& \textbf{FedCoLLM}& 27.4 & 36.0 & 45.4\\ 

 \hline

\multirow{3}*{ARC-E}& Zero-Shot& 53.2 & 65.6 &69.3\\

\cline{2-5}
~& Centralized&59.5 & 70.2 &76.8\\
 \cline{2-5}
~& \textbf{FedCoLLM}& 59.5  & 69.3 &75.2\\
 \hline

\end{tabular}

\caption{
We evaluate the server-side LLM performance of FedCoLLM using three experimental settings. Setting 1 (denoted as S1) deploys one server-side GPT-2-Large LLM paired with four client-side GPT-2-Small SLMs; Setting 2 (denoted as S2) uses one server-side OPT-6.7B LLM and four client-side OPT-1.3B SLMs; and Setting 3 (denoted as S3) adopts one server-side LLaMa2-7B LLM with four client-side LLaMa2-1.3B SLMs.
}

\label{tab:Homogeneous-LLM}
\end{table}

\begin{table}[!ht]
\centering
\footnotesize
\setlength{\tabcolsep}{2.8pt}

\begin{tabular}{c|c|c|c|c}
\hline
\multirow{2}{*}{\shortstack{\textbf{Task}}} & \multirow{2}{*}{\shortstack{{\textbf{Method}}}} & 
\multirow{2}{*}{\shortstack{S1: Clients \\ \textbf{GPT-2-Small}}} & 
\multirow{2}{*}{\shortstack{S2: Clients \\ \textbf{OPT-1.3B}}}  &
\multirow{2}{*}{\shortstack{S3: Clients \\ \textbf{LLaMa2-1.3B}}}\\
~ & ~ & ~ & ~& ~ \\  

\hline

\multirow{4}*{CQA}& Zero-Shot& 28.3 & 41.9 & 30.1 \\

\cline{2-5}
~& Standalone& 40.5 & 57.3 &56.1\\
\cline{2-5}
~& FedAvg& 41.7 &58.6 &56.8\\
 \cline{2-5}
~& \textbf{FedCoLLM}& \textbf{43.0} & \textbf{60.8} &\textbf{57.7}\\

 \hline

 \multirow{4}*{OBQA}& Zero-Shot& 16.4 & 23.4 & 23.2\\

\cline{2-5}
~& Standalone& 17.3 &26.7 &27.8\\
\cline{2-5}
~& FedAvg& 15.8 & 27.2 &26\\
 \cline{2-5}
~& \textbf{FedCoLLM}& \textbf{17.8} & \textbf{29.2} &\textbf{28.8}\\

 \hline
 
\multirow{4}*{ARC-C}& Zero-Shot& 19.0 & 23.4 &26.7\\

\cline{2-5}
~& Standalone& 21.4 & 28.2 &29.3\\

\cline{2-5}
~& FedAvg& 20.9 & 28.8 &29.4\\
 \cline{2-5}
~& \textbf{FedCoLLM}& \textbf{21.8} & \textbf{29.4} &\textbf{30.6}\\

 \hline
 
\multirow{4}*{ARC-E}& Zero-Shot& 43.8 & 57.0  &53.1\\

\cline{2-5}
~& Standalone& 46.4 & 59.93 &60.1\\

\cline{2-5}
~& FedAvg& 46.3 & 59.97 &60.3\\
 \cline{2-5}
~& \textbf{FedCoLLM}& \textbf{46.8} & \textbf{60.01} & \textbf{61.1}\\

 \hline
 \hline

\end{tabular}

\caption{
We evaluate the client-side SLM performance of FedCoLLM across three experimental settings. The results reported in the table are the average of all clients.
}
\label{tab:Homogeneous-SLM}
\end{table}

\subsection{Communication Cost}

We investigated the communication cost of FedCoLLM with LoRA, focusing on fine-tuned parameters. As shown in Table \ref{tab:model_size}, FedCoLLM significantly reduces communication costs: it only incurs 0.29\% of GPT-2's, 0.24\% of OPT's, and 0.23\% of LLaMa2's costs when fine-tuning all parameters.

\begin{table}[!h]
\centering
\footnotesize
\begin{tabular}{c|c|c|c}
\hline
\hline
\multicolumn{1}{c}{Model} & \multicolumn{1}{c}{Method} & \multicolumn{1}{c}{Transmitted Param Size (M)} & \multicolumn{1}{c}{Param Percent (\%)}  \\
\hline
\hline
 
\multirow{2}*{GPT-2}   & Full Model    &  124&100\\
\cline{2-4}
~         & \textbf{FedCoLLM}  &       \textbf{0.29 }&\textbf{0.24}\\ 
\hline

\multirow{2}*{OPT}   & Full Model    &  1316&100\\
\cline{2-4}
~         & \textbf{FedCoLLM}  &       \textbf{3.15}&\textbf{0.24}\\ 

\hline
\multirow{2}*{LLaMa2}   & Full Model    & 1345 & 100\\
\cline{2-4}
~         & \textbf{FedCoLLM}  &         \textbf{3.15}   &  \textbf{0.23}\\ 

\hline
\hline
\end{tabular}
\caption{Comparison of communication cost for FedCoLLM fine-tuning all parameters, fine-tuning using LoRA.}
\label{tab:model_size}
\end{table}

\section{Conclusions}
We propose FedCoLLM, an innovative and parameter-efficient federated co-tuning framework for LLMs and SLMs. This framework is designed to seamlessly adapt LLMs to resource-constrained downstream enterprises while preserving privacy, eliminating the need to deploy LLMs directly within these enterprises. FedCoLLM achieves this by introducing a SLM that acts as a bridge between the private data held by clients and the LLM model hosted on the server. Through the FedCoLLM training process, we obtain an LLM enriched with knowledge from multiple domains and a high-performing client SLMs, guided by the LLM.


\bibliographystyle{splncs04}
\bibliography{fedcollm}

\end{document}